# Improving Generalizability of Hip Fracture Risk Prediction via Domain Adaptation Across Multiple Cohorts

**AUTHORS**

Shuo Sun[1], Meiling Zhou[2], Chen Zhao[3], Joyce H. Keyak[4], Nancy E. Lane[5], Jeffrey D. Deng[6], Kuan-Jui Su[7], Hui Shen[7], Hong-Wen Deng[7], Kui Zhang[1,*], Weihua Zhou[8,9,*]

[1] Department of Mathematical Sciences, Michigan Technological University, Houghton, MI, USA

[2] Department of Statistics, Grand Valley State University, Allendale, MI, USA

[3] Department of Computer Science, Kennesaw State University, Marietta, GA, USA

[4] Department of Radiological Sciences, Department of Biomedical Engineering, and Department of Mechanical and Aerospace Engineering, University of California, Irvine, CA, USA

[5] Department of Internal Medicine and Division of Rheumatology, UC Davis Health, Sacramento, CA, USA

[6] Geisel School of Medicine, Dartmouth College, Hanover, NH, USA

[7] Division of Biomedical Informatics and Genomics, Tulane Center of Biomedical Informatics and Genomics, Deming Department of Medicine, Tulane University, New Orleans, LA, USA

[8] Department of Applied Computing, Michigan Technological University, Houghton, MI, USA

[9] Center for Biocomputing and Digital Health, Institute of Computing and Cybersystems, and Health Research Institute, Michigan Technological University, Houghton, MI, USA

* Corresponding authors:

Weihua Zhou, Ph.D.

E-Mail: whzhou@mtu.edu

Department of Applied Computing, Michigan Technological University, Houghton, MI, USA

Kui Zhang, Ph.D.

Department of Mathematical Sciences, Michigan Technological University, Houghton, MI, USA

E-Mail: kuiz@mtu.edu

**Abstract**

Clinical risk prediction models often fail to be generalized across cohorts because underlying data distributions differ by clinical site, region, demographics, and measurement protocols. This limitation is particularly pronounced in hip fracture risk prediction, where the performance of models trained on one cohort (the source cohort) can degrade substantially when deployed in other cohorts (target cohorts).

We used a shared set of clinical and DXA-derived features across three large cohorts - the Study of Osteoporotic Fractures (SOF), the Osteoporotic Fractures in Men Study (MrOS), and the UK Biobank (UKB), to systematically evaluate the performance of three domain adaptation methods - Maximum Mean Discrepancy (MMD), Correlation Alignment (CORAL), and Domain - Adversarial Neural Networks (DANN) and their combinations. For a source cohort with males only and a source cohort with females only, domain-adaptation methods consistently showed improved performance than the no-adaptation baseline (source-only training), and the use of combinations of multiple domain adaptation methods delivered the largest and most stable gains. The method that combines MMD, CORAL, and DANN achieved the highest discrimination with the area under curve (AUC) of 0.88 for a source cohort with males only and 0.95 for a source cohort with females only), demonstrating that integrating multiple domain adaptation methods could produce feature representations that are less sensitive to dataset differences. Unlike existing methods that rely heavily on supervised tuning or assume known outcomes of samples in target cohorts, our outcome-free approaches enable the model selection under realistic deployment conditions and improve generalization of models in hip fracture risk prediction.

## Introduction

Osteoporotic hip fractures represent a major global health concern, contributing substantially to mortality, disability, and long-term healthcare burden among aging populations[1-3]. Accurate identification of individuals at high risk for hip fractures is critical for timely prevention and early intervention. However, existing clinical tools such as bone mineral density (BMD) T-scores and the FRAX algorithm exhibit limited sensitivity - particularly among individuals without prior fracture history or those with BMD values in borderline osteogenic ranges[4-7]. Although recent advances in machine learning methods have demonstrated improved discriminative performance for hip fracture risk prediction, generalization of those models trained from a cohort to other cohorts remains a significant challenge[8-11] in real world clinical settings.

A central challenge underlying this limitation is the domain shift – a result from systematic differences in demographic composition, measurement practices, clinical settings, and population health characteristics across cohorts[12-14]. Models trained on one cohort (the source cohort) often exhibit substantial performance degradation when they are applied to other cohorts (target cohorts), even when harmonized features are used[15, 16]. In addition, most existing studies use outcomes of samples in target cohorts for cross-validation or parameter tuning – an assumption that is rarely feasible in real-world clinical deployment senerios[17].

Domain adaptation provides a possible solution to this challenge by aligning underlying data distributions between the source cohort and target cohorts. Classical domain adaptation approaches such as Maximum Mean Discrepancy (MMD) based, Correlation Alignment (CORAL) based, and Adversarial Neural Networks (DANN) based methods have shown strong performance in computer vision and natural language processing[18-21]. However, their application to clinical risk prediction, especially in situations when outcomes of samples in target cohorts are unavailable, remains limited[22-24]. More importantly, nearly all existing methods use target-cohort outcomes during hyperparameter selection, potentially leaking target-outcome information and inflating reported performance (e.g., model selection by target risk using labeled target hold-out data is considered ‘controversial’ in unsupervised domain adaptation, and source-risk selection is biased under domain shift)[28].

To overcome these limitations and assess the applicability of domain adaptation methods to hip fracture risk prediction, we proposed a novel hyperparameter selection strategy based on mean distributional discrepancy, enabling model selection without knowing outcomes of samples in target cohorts. Using three large cohorts - the Study of Osteoporotic Fractures (SOF), the Osteoporotic Fractures in Men Study (MrOS), and the UK Biobank (UKB), we systematically evaluated the performance of three widely used domain adaptation approaches – MMD based, CORAL based, and DANN based methods as well as their combinations in hip fracture risk prediction.

## Methods

### Datasets

We used three datasets: SOF, MrOS, and UKB. SOF and MrOS are prospective longitudinal studies of older females and males, respectively, in which incident fractures were systematically adjudicated over a 10–15-year follow-up. SOF is a prospective cohort of older women and originally enrolled 9,704 community-dwelling females, and MrOS is a prospective cohort of older men and originally enrolled 5,994 males.  In our analyses, SOF (females) and MrOS (males) served as the source cohorts for model development. Specifically, we selected SOF samples who (i) had a defined incident hip-fracture outcome over follow-up, and (ii) had complete-case availability for all predictors in the shared harmonized feature set used for domain adaptation (i.e., the same clinical and DXA-derived variables required in SOF/MrOS/UKB after recoding). Applying these inclusion criteria yielded approximately 3,625 SOF females in the source cohort (training cohort). Similarly, we selected MrOS samples who (i) had a defined incident hip-fracture outcome over follow-up, and (ii) had complete-case availability for the same shared predictor set after harmonization, yielding approximately 4,295 MrOS males in this source cohort. The reduction from the number of sample in the original enrolled cohorts to the number of samples in the source cohorts is expected because domain adaptation requires a strict shared predictor space across cohorts; therefore, participants with missing values in any required predictor or with non-harmonizable/ambiguous categorical responses were excluded to ensure consistent feature definitions and units across cohorts. From these two cohorts, we identified 3,625 females (SOF) and 4,295 males (MrOS) and used them as the source cohorts for model development. The target cohorts were constructed from UKB, a contemporary population-based cohort with 502,128 samples. To selected samples in the target cohorts suitable for our evaluation, we applied complete-case selection over the shared predictors and excluded ambiguous response categories. Specifically, samples were retained only if all harmonized clinical predictors and DXA-derived BMD measures required by the model were non-missing. In addition, for categorical predictors we removed non-informative or ambiguous response categories (e.g., “Prefer not to answer”, “Do not know”, and “None of the above”), because these responses cannot be reliably mapped to the unified coding scheme shared across cohorts. These steps ensured that all included UKB samples were represented in the same fixed feature space as the source cohorts, enabling valid unsupervised alignment and external evaluation. The selection from  UKBwas performed as follows. We first extracted the Instance-2 measurements corresponding to the shared predictor set (age, sex, height, weight, left/right grip strength, usual walking pace, smoking status, difficulty walking long distances, and DXA-derived femoral neck/total hip/lumbar spine BMD measures). We then removed non-informative or ambiguous response categories that cannot be reliably harmonized across cohorts (walking pace: “None of the above”; functional difficulty: “Do not know”; smoking: “Prefer not to answer”). Grip strength was defined as the mean of left- and right-hand measurements. Smoking status was encoded as Never=0, Former=1, and Current=2, and functional difficulty was encoded on an ordinal 0–4 scale. Hip fracture status within 10 years of follow-up was defined from hospital inpatient records (binary outcome-label with 1 indicating incident hip fracture). To improve comparability of DXA measures, BMD values were converted to T-score–like standardized variables within UKB using fracture-free participants as the reference within two age strata (65–75 and 76–96). Finally, we applied complete-case selection requiring non-missing values for all predictors used in the domain adaptation models. After these steps, 410 females (5 hip-fracture cases) and 210 males (3 hip-fracture cases) were selected and were used as a target cohort with females only and a target cohort with males only. Each target cohort was randomly split into a pseudo-training cohort and a held-out evaluation cohort with the same number of samples and were stratified by hip-fracture status. The pseudo-training cohort was used as the target cohort to compute domain-alignment losses, and the outcomes of those samples were not used for training, tuning, or model selection. All reported performance metrics were computed exclusively on the held-out evaluation cohort.

The outcome used in our paper was the incident hip fracture within a 10-year follow-up window and is defined by using adjudicated outcomes in SOF/MrOS and hospital inpatient records in UKB. All analyses were restricted to a shared predictor set consisting of nine clinical predictors (age, height, weight, grip strength, usual walking pace, smoking status, two binary indicators for prior fracture history -(upper arm/shoulder fracture and wrist fracture], and an ordinal functional-mobility/IADL-related walking-difficulty score) and three DXA-derived BMD predictors (femoral neck, total hip, and lumbar spine BMD expressed as T-score–like values). These predictors were selected based on cross-dataset availability and clinical relevance to hip fracture risk.

**Data notations and model architectures**

Let $X_s$ be a $n_s$ by $d$ matrix and $y_s$ be a length $n_s$ vector that denote the harmonized feature matrix and the binary hip fracture outcomes from the training dataset (the source cohort), respectively. Here $n_s$ is the number samples in the source cohort, $d$ is the number of predictors, and a 1 in $y_s$ indicates an incident hip fracture and $a$ 0 in $y_s$ indicates no hip fracture, respectively. Let $X_t$be a $n_t$ by $d$ matrix and $y_t$ be a length $n_t$ vector that denote the harmonized feature matrix and the binary hip fracture outcomes from the external validation dataset (the target cohort), respectively. Here $n_t$ is the number of samples in the target cohort. Note that $y_t$ is assumed to be unavailable and is not used during model training and hyperparameter selection and $y_t$ is used only in evaluating the model performance. Here $d = 12$ harmonized predictors were used.

We used a feature extractor $G(\cdot;\theta_g)$ to map each harmonized feature vector (a length $d$ vector which is a row of the harmonized feature matrix) into an embedding representation vector with the length of $p$, followed by a classifier $C(\cdot;\theta_c)$ that outputs a hip fracture probability. Specifically, we have

$$h = G(x;\theta_g), \qquad \hat{y} = C(h;\theta_c).$$

Here, $x$ is a length $d$ vector which is a row of the harmonized feature matrix and represents the harmonized feature of a sample, $h$ is a length $p$ vector, and $\hat{y}$ is the predicted probability for the hip fracture risk. The embedding representation $h$ is a learned latent feature vector produced by the feature extractor $G(\cdot;\theta_g)$. It is an intermediate representation that summarizes the original $d$-dimensional vector into a $p$-dimensional space, where samples with similar clinical profiles are encouraged to have similar representations. The feature extractor is a lightweight two-layer Multiple Layer Perceptron (MLP) with Rectified Linear Unit (ReLU) nonlinearities (and regularization such as normalization/dropout in implementation), and the classifier head is a linear logit layer followed by a sigmoid transform. The feature extractor $G(\cdot;\theta_g)$, classifier $C(\cdot;\theta_c)$, and (when adversarial alignment is used) domain discriminator $D(\cdot;\theta_d)$ are neural networks parameterized by trainable weights and biases. Here, $\theta_g$ denotes the collection of all trainable parameters in the feature extractor G, including the weight matrices and bias vectors of its linear layers (e.g., $\{W_1, b_1, W_2, b_2\}$) and the affine parameters of normalization layers when present (e.g., LayerNorm or BatchNorm scale and shift parameters). In our implementation, normalization layers are indeed used: the female pipeline includes LayerNorm layers inside the feature extractor, and the male pipeline includes BatchNorm layers. The prediction head C is a single linear logit layer followed by a sigmoid transform, so $\theta_c = \{w, b\}$ denotes its weight vector w and bias b. When adversarial alignment is used, $\theta_d$ denotes all trainable parameters in the domain discriminator D (i.e., the weights and biases of its layers), which is trained to predict a domain indicator (source vs. target) from the learned embeddings.

All trainable parameters $(\theta_g, \theta_c, \theta_d)$ are initialized with standard random initialization (e.g., PyTorch default initialization) and estimated via mini-batch stochastic gradient descent (AdamW optimizer in our implementation) by minimizing a composite objective that combines the source supervised task loss and one or more unsupervised alignment losses. Concretely, $\theta_g$ and $\theta_c$ are updated to improve hip-fracture prediction on the outcome-labeled source data through $\mathcal{L}_{task}$. When adversarial domain adaptation is enabled, $\theta_d$ is updated to distinguish source vs. target embeddings by minimizing the domain classification loss $\mathcal{L}_{dom}$, while $\theta_g$ is simultaneously encouraged (via a gradient reversal layer) to produce domain-invariant embeddings that make this discrimination difficult. Thus, $\theta_d$ is optimized to separate domains, whereas $\theta_g$ is optimized both for prediction and for alignment, under the same training configuration.

The embedding dimension $p$ is the output width of the feature extractor $G(\cdot;\theta_g)$. It is a model hyperparameter that controls representation capacity and is selected during model configuration (e.g., via our outcome label-free hyperparameter selection procedure described below). In the final experiments, we used $p = 256$ consistently across female and male pipelines to avoid introducing sex-specific architectural differences; sensitivity analyses over alternative $p$ values are reported elsewhere (if applicable).

Because hip fractures are rare, training on the source cohort incorporated some strategies to handle such imbalance, including weighted sampling in mini-batches and a mild positive-class augmentation fallback to stabilize training in extremely imbalanced splits.

Let $(x_{s,i}, y_{s,i})$ denote the harmonized predictor and the outcome for the $i$-th source sample, where $x_{s,i}$ is the harmonized predictor vector with the length of $d$ and $y_{s,i}$ indicates incident hip fracture (here $y_{s,i}$ is either 0 or 1 and 1 indicates the hip fracture). The feature extractor and classifier produce a probability $\hat{y}_{s,i} = C\left(G\left(x_{s,i}; \theta_g\right); \theta_c\right)$ for the hip fracture risk. Model parameters $\left(\theta_g, \theta_c\right)$ were estimated by minimizing a (class-weighted) binary cross-entropy objective:

$$\mathcal{L}_{\text{task}}\left(\theta_g, \theta_c\right) = \frac{1}{n_s} \sum_{i=1}^{n_s} L\left(\hat{y}_{s.i}, y_{s,i}\right),$$

where $L(\hat{y}, y)$ denotes the weighted binary cross-entropy,

$$L(\hat{y}, y) = -\omega\, y \log(\hat{y}) - (1 - y) \log(1 - \hat{y}),$$

where $\omega = \frac{\text{number of samples with the hip fracture}}{\text{number of samples withou the hip fracture}} > 0$ is used to up-weight the number samples with the hip fracture to address imbalance between the number of samples with and without the hip fracture.

**Domain adaptation methods**

Three domain adaptation methods, the method based on the Maximum Mean Discrepancy (MMD), the method based on Correlation Alignment (CORAL), and the method based on Domain - Adversarial Neural Networks (DANN), and their combinations were evaluated.

**Maximum Mean Discrepancy – MMD based methods**

The MMD based methods reduce distributional mismatch between source and target embeddings by minimizing the Maximum Mean Discrepancy (MMD), a nonparametric distance between two distributions defined in a reproducing kernel Hilbert space (RKHS). In our setting, we used a Gaussian radial basis function (RBF) kernel. Specifically, the Gaussian RBF kernel for two vectors $h_1$ and $h_2$ are defined as:

$$k(h, h') = exp\left(-\frac{||h_2 - h_2||_2^2}{2\sigma^2}\right),$$

where $||\cdot||_2$ is the Euclidean norm and $\sigma > 0$ is the kernel bandwidth controlling the scale of similarity. In practice, the kernel bandwidth is not fixed. We used a median heuristic on mini-batch embeddings and a multi-scale RBF kernel: we set $\sigma^2$ proportional to the median of all the pairwise squared distance between an embedding representation vector from the source cohort and an embedding representation vector from the target cohort and used the average of MMD over possible values of $\sigma^2$. Here, we used $\sigma^2 \in \{0.5 \cdot \text{"base"}, 1.0 \cdot \text{"base"}, 2.0 \cdot \text{"base"}\}$. Here, base is the mini-batch median of pairwise squared distances in the embedding space (computed from the concatenated source target embeddings); it varies across batches and training epochs, with a fallback of 1.0 in degenerate cases.

The standard empirical estimator of MMD is:

$$\mathcal{L}_{\mathcal{MMD}} = \frac{1}{n_s^2}\sum_{i=1}^{n_s}\sum_{i'=1}^{n_s} k\left(h_{s,i}, h_{s,i'}\right) + \frac{1}{n_t^2}\sum_{j=1}^{n_t}\sum_{j'=1}^{n_t} k\left(h_{t,j}, h_{t,j'}\right) - \frac{2}{n_s n_t}\sum_{i=1}^{n_s}\sum_{j=1}^{n_t} k\left(h_{s,i}, h_{t,j}\right).$$

Here, $h_{s,i}$ $(i = 1, \cdots, n_s)$ is the embedding representation vector for the $i$-th sample in the source cohort and $h_{t,j}$ $(j = 1, \cdots, n_t)$ is the embedding representation vector for the $j$-th sample in the target cohort. Minimizing $\mathcal{L}_{\mathcal{MMD}}$ was achieved by adding it to the training objective and updating the feature extractor parameters via AdamW. At each iteration, $\mathcal{L}_{\mathcal{MMD}}$ was computed on mini-batch embeddings from the samples in the source cohort and in the target cohort gradients were backpropagated through $G(\cdot;\theta_g)$.

**Correlation Alignment (CORAL) based methods**

The second mechanism matches second-order structure using correlation alignment (CORAL). Let $\Sigma_s$ and $\Sigma_t$ denote the empirical covariance matrices of source and target embeddings, respectively. Specifically, $\Sigma_s = \frac{1}{n_s - 1}\sum_{i=1}^{n_s}\left(h_{s,i} - \bar{h}_s\right)\left(h_{s,i} - \bar{h}_s\right)^{\top}$, $\Sigma_t = \frac{1}{n_t - 1}\sum_{j=1}^{n_t}\left(h_{t,j} - \bar{h}_t\right)\left(h_{t,j} - \bar{h}_t\right)^{\top}$, where $\bar{h}_s$ and $\bar{h}_t$ are the mean of the embedding representation vectors of the source and target cohorts, respectively. CORAL minimizes their discrepancy:

$$\mathcal{L}_{\mathcal{CORAL}} = \frac{1}{4p^2}||\Sigma_s - \Sigma_t||_F^2$$

thereby encouraging source and target embeddings to share similar variance–covariance patterns. In implementation, we used a generalized CORAL variant that replaces the Frobenius norm with an $l_q$-norm penalty; the Frobenius norm corresponds to the special case $q = 2$.

**Domain - Adversarial Neural Networks (DANN) based methods**

The third mechanism adopts adversarial domain confusion. A domain discriminator $D(\cdot;\theta_d)$ predicts whether an embedding originated from the source (training) or target (validation) domain. Domain labels (domain indicators) are defined as $d = 0$ for a sample from the source cohort and $d = 1$ for a sample from the target cohort. The discriminator minimizes the binary cross-entropy loss

$$\mathcal{L}_{dom} = -\frac{1}{n_s + n_t} \sum_{i=1}^{n_s+n_t} d_i \log\left(\widehat{d_i}\right) + (1 - d_i) \log\left(1 - \widehat{d_i}\right),$$

where $\widehat{d_i} = D(h_i; \theta_d)$is the predicted probability that $h_i$ comes from the target domain.

To encourage domain-invariant embeddings, gradients flowing from the discriminator to the feature extractor are reversed using a gradient reversal layer (GRL). The GRL is the identity in the forward pass but multiplies the backpropagated gradient by $-\lambda$ before it reaches the feature extractor, forcing $G(\cdot;\theta_g)$ to learn representations that confuse $D(\cdot;\theta_d)$.

When all three alignment modules are enabled, the complete objective is

$$\mathcal{L} = \mathcal{L}_{\text{task}} + \lambda_{\text{MMD}}\,\mathcal{L}_{MMD} + \lambda_{\text{CORAL}}\,\mathcal{L}_{CORAL} + \lambda_{\text{GRL}}\,\mathcal{L}_{dom}.$$

where $\lambda_{\text{MMD}}, \lambda_{\text{CORAL}}, \lambda_{\text{GRL}} \geq 0$ control the strength of each alignment mechanism. $\mathcal{L}_{\text{task}}$ is the only term that uses outcomes, and it directly trains the model to be predictive of hip fracture risk on the source cohort. In the joint objective, $\mathcal{L}_{\text{task}}$ anchors the representation to remain outcome-informative, while the alignment losses $\mathcal{L}_{MMD}, \mathcal{L}_{CORAL}, \mathcal{L}_{dom}$ encourage the learned embeddings to be transferable across domains. The coefficients $\lambda_{\text{MMD}}, \lambda_{\text{CORAL}}, \lambda_{\text{GRL}}$ control the trade-off between supervised learning on the source cohort and distribution alignment between the source and target cohorts. To avoid any use of outcomes from the samples of target cohorts, these coefficients were treated as preset hyperparameters and assigned using a simple module-count heuristic: smaller weights were used when only one or two alignment modules were enabled, and larger weights were used when all alignment modules were enabled. When MMD and CORAL were simultaneously activated (our main setting), we set $\lambda_{\text{MMD}} = \lambda_{\text{CORAL}} = 0.7$ in both female and male experiments. For the adversarial component (DANN with a gradient reversal layer, GRL), we used a fixed domain-loss weight together with a monotone GRL schedule that increases over training epochs. The domain-loss weight was set to 0.7 for cohorts with females only and 0.8 for cohorts with males only to account for differences in domain shift magnitude and class imbalance between genders.

**Outcome-free hyperparameter selection**

One of our contributions is the development of an outcome-free hyperparameter selection criterion that does not rely on outcomes of the samples on target cohorts. Let $\eta$ denote a hyperparameter configuration (e.g., learning rate, weight decay, batch size, hidden width/embedding dimension, and alignment weights). For each $\eta$, we train the model on the source cohort in which the outcomes are known (and the target cohort in which the outcomes are not used) to obtain fitted parameters $\theta_g^*(\eta), \theta_c^*(\eta), \theta_d^*(\eta)$. Based on these hyperparameters, we obtained the embedding representation vectors and then MMD (in other words, $\mathcal{L}_{\mathcal{MMD}}$). .

The set of hyperparameter that minimizes MMD was used. Thus, the hyperparameters used produce the most closely aligned source-target representations. Note that $MMD$ is computed from the learned embeddings and therefore depends primarily on $\theta_g^*(\eta)$; however, $\theta_c^*(\eta)$ and $\theta_d^*(\eta)$ are jointly learned under the same training configuration $\eta$ and influence $\theta_g^*(\eta)$ through backpropagation. Thus, selecting $\eta$implicitly selects the full set of trained parameters$\theta_g, \theta_c$, and $\theta_d$.

**Role of the UKB pseudo-training split and evaluation set**

Across all evaluations, target-domain hip fracture outcomes were not used for model tuning, thresholding, or hyperparameter selection. For each sex-specific experiment, the UKB target subset was randomly split 50/50 into (i) a pseudo-training split and (ii) a held-out test split using stratification on the hip fracture outcome. The only role of the UKB pseudo-training split was to provide target samples without outcomes used to compute domain-alignment losses (e.g., MMD/CORAL computed between source vs. target embeddings, and the domain-discriminator loss in DANN). Importantly, outcomes of the UKB samples were not used for training or model selection in the domain adaptation setting.

Hyperparameters were selected using the proposed outcome-free criterion $\Delta(\eta) = \mathrm{MMD}^2\big(H_s(\eta), H_t(\eta)\big)$ via a grid search over standard optimization and architecture settings, including learning rate $\{5 \times 10^{-4}, 10^{-3}, 2 \times 10^{-3}\}$, weight decay $\{10^{-5}, 10^{-4}, 5 \times 10^{-4}\}$, batch size $\{64,128,256\}$, and hidden/embedding size $\{128,256,512\}$. In the female experiment, we performed outcome-free hyperparameter selection by grid search and selected a configuration with learning rate $10^{-3}$, weight decay $10^{-5}$, batch size 64, and hidden/embedding size 256. In the male experiment, we used the same model family and the same outcome-free selection principle, but the training script fixed the batch size at 128 for stability, while tuning other optimization/architecture hyperparameters (e.g., learning rate, weight decay, and hidden/embedding size). Consistent with this design, the adversarial component (DANN/GRL) was also assigned a slightly stronger domain-loss weight in the male setting (0.8 vs. 0.7 in females), reflecting the empirically larger source–target discrepancy and the increased sensitivity of training under rarer events. Training used AdamW with early stopping based on a held-out source validation split and gradient clipping for stability.

All performance metrics were computed only on the held-out UKB test split, which remained untouched during training and alignment. Thus, the pseudo-training split influences the model only through alignment losses, whereas the reported AUC/ΔAUC values are obtained exclusively from the held-out UKB test split. The achieved improvements therefore reflect genuine distributional alignment rather than indirect leakage of predictive information. Altogether, these results show that integrating unsupervised domain adaptation—particularly when combining MMD, CORAL, and DANN in a unified framework—leads to consistent enhancements in fracture-risk prediction performance.

## Results

### Harmonized feature distributions and preprocessing.

Sample size and summary statistics for 12 harmonized predictors are presented in **Table 1** (SOF females: n=3,625; MrOS males: n=4,295; UKB females: n=410; UKB males: n=210). Several systematic between-cohort differences remain evident, indicating meaningful domain shift between the source cohorts and the sex-matched UKB target subsets.

First, UKB participants are substantially younger than the source cohorts. The mean age is 71.5 ± 5.3 years in SOF and 74.1 ± 6.0 years in MrOS, whereas the mean ages in UKB females and UKB males are 60.0 ± 4.1 and 60.7 ± 4.0 years, respectively. This ~11–14-year age gap reflects a pronounced demographic shift that may alter baseline fracture risk and the distributions of correlated predictors.

Second, anthropometric and strength measures show consistent sex patterns across cohorts. Males are taller and heavier than females in both the source and target cohorts (height: 173.9 ± 6.8 cm in MrOS and 176.8 ± 7.0 cm in UKB males vs. 159.1 ± 5.9 cm in SOF and 162.9 ± 6.3 cm in UKB females; weight: 82.6 ± 13.1 kg in MrOS and 81.6 ± 12.3 kg in UKB males vs. 67.8 ± 12.3 kg in SOF and 67.0 ± 12.8 kg in UKB females). Grip strength follows the same pattern (SOF: 20.9 ± 4.2 kg; UKB females: 20.9 ± 5.5 kg; MrOS: 38.7 ± 8.3 kg; UKB males: 35.5 ± 9.2 kg).

Third, the DXA-derived BMD measures (T-scores) show notable source–target shifts, particularly in UKB. By construction, the source cohorts are centered at 0.00 with SD near 1 for total hip, lumbar spine, and femoral neck T-scores (0.00 ± 1.01, 0.00 ± 1.00, 0.00 ± 1.01 in SOF; 0.00 ± 1.01, 0.00 ± 1.00, 0.00 ± 1.01 in MrOS). In contrast, UKB females have lower mean T-scores (total hip −0.48 ± 0.72; lumbar spine −0.40 ± 0.76; femoral neck −0.26 ± 0.86), while UKB males have higher mean T-scores (total hip 0.91 ± 0.80; lumbar spine 0.77 ± 0.95; femoral neck 0.53 ± 1.02). These shifts indicate residual differences in bone density distributions between the source and UKB target domains even after standardization.

Finally, the additional four predictors reported as counts/percentages also demonstrate clear distribution differences across cohorts. For usual walking pace, UKB shows substantially fewer "Slow pace" participants than SOF/MrOS (Slow pace: SOF 917 (25.3%), MrOS 1,078 (25.1%), UKB females 31 (7.6%), UKB males 19 (9.0%)) and a much larger proportion reporting "Brisk pace" (UKB females 173 (42.2%), UKB males 81 (38.6%) vs. SOF 863 (23.8%) and MrOS 1,067 (24.8%)). For smoking status, MrOS contains a high proportion of former smokers (2,490 (58.0%)) compared with UKB males (82 (39.0%)) and UKB females (161 (39.3%)), while current smoking is relatively uncommon in all cohorts (SOF 285 (7.9%), MrOS 130 (3.0%), UKB females 6 (1.5%), UKB males 7 (3.3%)). Prior fracture history also differs: upper arm/shoulder fracture is more common in UKB (UKB females 50 (12.2%), UKB males 22 (10.5%)) than in MrOS (50 (1.2%)) and is intermediate in SOF (219 (6.0%)); wrist fracture is especially frequent in UKB females (107 (26.1%)) compared with SOF (393 (10.8%)) and MrOS (69 (1.6%)), with UKB males at 24 (11.4%). Together, these categorical shifts—alongside the continuous-feature differences—support the presence of substantive domain shift between SOF/MrOS and UKB, motivating the use of domain adaptation methods to improve external generalization.

Results from the methods without and with different domain adaptation approaches are presented in **Tables 2 and 3**. In can be seen that across both cohorts, domain adaptation methods consistently improved discrimination on the held-out UKB evaluation cohort compared with the baseline method without the use of domain adaptation. Although alignment uses no target outcomes, it still uses target features; therefore, we held out half of the target cohort to avoid evaluating on the same target individuals used to drive the alignment objective, yielding an unbiased estimate of generalization to unseen target participants. For the cohorts with the female samples only, the source cohort is SOF. The method (baseline method) without any domain adaptation approaches yielded a mean area under curve (AUC) near 0.80. The methods with a single domain adaptation approach resulted in larger AUCs, but the magnitude of increasement depend on which domain adaptation method was used: comparing with the AUC from the baseline method, the method with MMD and the method with DAN achieved a mean AUC increase of ~0.14 and ~0.13, respectively. Both improvements were highly significant ($p < 0.001$) and represent the largest gains in terms of AUC for methods with a single domain adaptation method. Methods with two or two or three domain adaptation approaches showed further improvements in terms of AUC. P-values were computed using a paired t-test across five random seeds, comparing the AUC of each domain-adaptation variant against the baseline AUC on the same held-out UKB test split, with the paired differences $\Delta^{(s)} = \mathrm{AUC}_{\mathrm{m}}^{(\mathrm{s})} - \mathrm{AUC}_{\mathrm{base}}^{(\mathrm{s})}$. Among all methods with at least one of three domain adaptation approaches the method with all three domain adaptation methods (MMD + CORAL + DANN) achieved the highest average AUC of 0.96 and had the notably higher average AUC than other six methods.

Results for the male cohort from **Table 3** showed a similar pattern, though with slightly lower AUCs, consistent with the much smaller proportion of samples with the hip fracture in the corresponding target cohort (UKB male samples, 3/210) and the substantial distribution shift between the source cohort (MrOS, 161/4,295) and the target cohort (UKB male samples). For the male experiment, because the number of UKB hip-fracture cases was extremely small, the UKB target subset could not support a perfectly balanced stratified 50/50 split. We therefore used a split that ensured at least one case in each partition (pseudo-training vs. evaluation). The pseudo-training partition was used only for alignment losses, and Table 3 reports performance on the held-out evaluation partition. The baseline method had an average AUC of 0.71. For three methods with a single domain adaptation method, MMD again yielded the largest improvement in terms of AUC, with a mean increase of 0.156 and $p < 0.001$. The method with DANN also significantly increased the AUC of the baseline method, although the increment was around 0.09 and slightly lower than the method with MMD. The method with CORAL alone did not perform significantly better than the baseline method, a similar pattern observed for the cohorts with female samples only. Again, methods with more domain adaptation approaches larger AUCs and the method with all three-domain adaptation approaches achieved the highest average AUC of 0.88.

**Discussion**

In this study, we proposed a domain adaptation approach to improve the hip fracture risk prediction across multiple cohorts and systematically evaluated its performance. Our evaluations were conducted using sample cohorts, harmonization rules, and preprocessing steps as in our previous two-stage hip fracture prediction pipeline[11]. This ensures direct comparability and validates that improvements arise from domain adaptation rather than changes in data curation. Second, performance was reported across multiple seeds, with statistical testing confirming that improvements were not attributable to random initialization. Third, our hyperparameter selection approach do not use outcomes of samples from the target cohort throughout model selection and training. This prevents optimistic bias and mirrors the constraints of real-world deployment where fracture outcomes may not be available in new cohorts for years.

Nonetheless, there are several limitations in our current study. The reliance on the split of target cohort, while necessary to produce a pseudo-training set for domain-adaptation and an untouched test set, halves the number of samples available for both purposes. A larger dedicated target cohort would likely further stabilize hyperparameter selection and potentially catch true benefits from domain adaptation approaches. Additionally, while three complementary domain-adaptation methods were systematically evaluated, many alternative strategies - such as conditional alignment, optimal transport, or contrastive adaptation - were not explored in this work but may offer additional gains. Finally, all models were based on harmonized tabular inputs. Incorporating richer

DXA imaging features or multimodal predictors may require redesigned adaptation modules or higher-capacity encoders.

Despite these limitations, the findings demonstrate that domain adaptation methods - when combined with appropriate hyperparameter selection - can improve the hip fracture risk prediction across multiple cohorts. These findings represent progress toward models that preserve accuracy across diverse populations and measurement settings and illustrate how domain adaptation approaches can be incorporated into clinical prediction pipelines.

## Conclusion

In this study, we introduced a domain adaptation framework designed to improve the hip fracture risk prediction cross multiple cohorts. We showed that the methods based on domain adaption approaches achieved consistent and statistically significant improvements in predictive performance across both male and female cohorts. The methods with more domain adaption approaches performed better and the methods with all three domain adaptation approaches discuss in this study produced the highest AUC in every setting, confirming the complementary nature of mean alignment, covariance alignment, and adversarial invariance. These findings showed that domain adaptation approached can effectively improve the hip fracture risk prediction across multiple cohorts.

**Data Availability**

The data used in this study cannot be shared with the public due to third-party use restrictions and patient confidentiality concerns. The three datasets utilized in this study were obtained from dbGaP and UK Biobank websites, the links are:

The Osteoporotic Fractures in Men (MrOS) (phs000373.v1.p1)

https://www.ncbi.nlm.nih.gov/projects/gap/cgi-bin/study.cgi?study_id=phs000373.v1.p1

The Study of Osteoporotic Fractures (SOF) (phs000510.v1.p1)

https://www.ncbi.nlm.nih.gov/projects/gap/cgi-bin/study.cgi?study_id=phs000373.v1.p1

The UK Biobank (application ID: 61915)

https://www.ukbiobank.ac.uk/enable-your-research/about-our-data

MrOS and SOF contains only the male and female samples, respectively, with data obtained under the Disease-Specific (Aging Related 1, RD) consent group.

**References**

1. Sing CW, et al. Global Epidemiology of Hip Fractures: Secular Trends in Incidence, Treatment, and Mortality. Journal of Bone and Mineral Research. 2023;38(8):1064–1075.
2. Dong Y, et al. What was the Epidemiology and Global Burden of Disease of Osteoporosis in 2019? Journal of Orthopaedic Translation. 2022;36:9–18.
3. Shen Y, et al. The Global Burden of Osteoporosis, Low Bone Mass, and Related Fractures: Results from the Global Burden of Disease Study 1990–2019. Frontiers in Endocrinology. 2022;13:882241.
4. Silverman SL, Calderon AD. The Utility and Limitations of FRAX: A US Perspective. Current Osteoporosis Reports. 2010;8(4):192–197.
5. Hillier TA, et al. WHO Absolute Fracture Risk Models (FRAX): Do Clinical Risk Factors Improve Fracture Prediction in Older Women? Journal of Bone and Mineral Research. 2011;26(8):1774–1782.
6. El Miedany Y. FRAX: re-adjust or re-think. Archives of Osteoporosis. 2020;15(1):170.

7. Ye C, et al. Adjusting FRAX Estimates of Fracture Probability Based on Vertebral Fracture Assessment. JAMA Network Open. 2023;6(8):e2326829.

8. Wu Y, et al. Predictive Value of Machine Learning on Fracture Risk in Patients With Osteoporosis: A Systematic Review and Meta-analysis. BMJ Open. 2023;13:e071430.

9. Aging and Disease Review Group. Prediction Models for Osteoporotic Fracture Risk: A Systematic Review and Critical Appraisal. Aging and Disease. 2022;13(4):1215–1238.

10. Rietz M, Brønd J, Möller S, Søndergaard J, Abrahamsen B, Rubin K. Introducing FREMML: a decision-support approach for automated identification of individuals at high imminent fracture risk. Archives of Osteoporosis. 2025;20(1).

11. Sun S, et al. An Advanced Two-Stage Model with High Sensitivity and Generalizability for Prediction of Hip Fracture Risk Using Multiple Datasets. arXiv: 2510.15179. 2025.

12. Badgeley MA, et al. Deep learning predicts hip fracture using confounding variables. npj Nature Medicine. 2019.

13. Finlayson SG, et al. The Clinician and Dataset Shift in Artificial Intelligence. New England Journal of Medicine. 2021;385(3):283–286.

14. Subbaswamy A, Saria S. From development to deployment: dataset shift, causality, and shift-stable models in health AI. Biostatistics. 2020;21:345–352.

15. Futoma J, Simons M, Panch T, Doshi-Velez F, Celi LA. The myth of generalisability in clinical research and machine learning in health care. The Lancet Digital Health. 2020. 2020;2:e489–e490.

16. Goetz L, Seedat N, Vandersluis R, van der Schaar M. Generalization—a key challenge for responsible AI in patient-facing clinical applications. npj Digital Medicine. 2024;7:126.

17. Geirhos R, et al. Shortcut learning in deep neural networks. Nature Machine Intelligence. 2020.

18. Gretton A, et al. A kernel two-sample test (MMD). JMLR. 2012.

19. Sun B, Saenko K. CORAL: Correlation Alignment for deep DA. ECCV Workshops. 2016.

20. Ganin Y, et al. Domain-Adversarial Neural Networks (DANN). JMLR. 2016.

21. Tzeng E, et al. Adversarial DA with joint distribution alignment. CVPR. 2017.

22. Purushotham S, Meng C, Che Z, Liu Y. Benchmarking deep learning models on large healthcare datasets. Journal of Biomedical Informatics. 2018;83:112–134.

23. Singh H, Mhasawade V, Chunara R. Generalizability challenges of mortality risk prediction models: A retrospective analysis on a multi-center database. PLOS Digital Health. 2022.

24. Ghassemi M, et al. A review of challenges and opportunities in machine learning for health. Nature Medicine. 2020.

25. Zhao H, et al. On learning domain-invariant representations. ICML (PMLR). 2019..

26. Hu D, et al. Towards Reliable Model Selection for Unsupervised Domain Adaptation. NeurIPS (Datasets & Benchmarks Track). 2024.

27. Pineau J, Vincent-Lamarre P, Sinha K, et al. Improving Reproducibility in Machine Learning Research. arXiv:2003.12206. 2020.

28. You et al., Towards Accurate Model Selection in Deep Unsupervised Domain Adaptation. ICML. 2019.

**Author Information**

These authors jointly supervised this work: Kui Zhang, Weihua Zhou.

**Affiliations**

**Department of Mathematical Sciences, Michigan Technological University, Houghton, MI, USA**

Shuo Sun, Kui Zhang*

**Department of Statistics, Grand Valley State University, Allendale, MI, USA**

Meiling Zhou

**Department of Computer Science, Kennesaw State University, Marietta, GA, USA**

Chen Zhao

**Department of Radiological Sciences, Department of Biomedical Engineering, and Department of Mechanical and Aerospace Engineering, University of California, Irvine, CA, USA**

Joyce H. Keyak

**Department of Internal Medicine and Division of Rheumatology, UC Davis Health, Sacramento, CA**

Nancy E. Lane

**Geisel School of Medicine, Dartmouth College, Hanover, NH, USA**

Jeffrey D. Deng

**Division of Biomedical Informatics and Genomics, Tulane Center of Biomedical Informatics and Genomics, Deming Department of Medicine, Tulane University, New Orleans, LA**

Kuan-Jui Su, Hui Shen & Hong-Wen Deng

**Department of Applied Computing, Michigan Technological University, Houghton, MI, USA**

**Center for Biocomputing and Digital Health, Institute of Computing and Cybersystems, and Health Research Institute, Michigan Technological University, Houghton, MI, USA**

Weihua Zhou*

**Corresponding authors**

Correspondence to: Kui Zhang and Weihua Zhou.

ORCID IDs:

Kui Zhang: [0000-0002-2441-2064]

Weihua Zhou: [0000-0002-6039-959X]

**Acknowledgements**

This research has been conducted using the UK Biobank Resource under application number [61915]. It was in part supported by grants from the National Institutes of Health, USA (U19AG055373, 1R15HL172198, and 1R15HL173852) and American Heart Association (#25AIREA1377168).

**Ethics declarations**

Competing interests

All authors declare that there are no conflicts of interest.

## Tables

**Table 1. Sample size and** summary statistics (mean ± SD for continuous and count/frequency for binary or ordinal) for harmonized features for source and target cohorts.

| **Feature** | **SOF (Females)** | **MrOS (Males)** | **UKB Female** | **UKB Male** |
|---|---|---|---|---|
| Sample Size, n | 3625 | 4295 | 410 | 210 |
| Age (years) | 71.5 ± 5.3 | 74.1 ± 6.0 | 60.0 ± 4.1 | 60.7 ± 4.0 |
| Height (cm) | 159.1 ± 5.9 | 173.9 ± 6.8 | 162.9 ± 6.3 | 176.8 ± 7.0 |
| Weight (kg) | 67.8 ± 12.3 | 82.6 ± 13.1 | 67.0 ± 12.8 | 81.6 ± 12.3 |
| Grip Strength (kg) | 20.9 ± 4.2 | 38.7 ± 8.3 | 20.9 ± 5.5 | 35.5 ± 9.2 |
| Total Hip BMD (T-score) | 0.00 ± 1.01 | 0.00 ± 1.01 | –0.48 ± 0.72 | 0.91 ± 0.80 |
| Lumbar Spine BMD (T-score) | 0.00 ± 1.00 | 0.00 ± 1.00 | –0.40 ± 0.76 | 0.77 ± 0.95 |
| Femoral Neck BMD (T-score) | 0.00 ± 1.01 | 0.00 ± 1.01 | –0.26 ± 0.86 | 0.53 ± 1.02 |
| IADL score (0–5) | 0.52 ± 0.96 | 0.34 ± 0.81 | 0.40 ± 0.84 | 0.32 ± 0.80 |
| Usual walking pace, % | | | | |
| (Slow pace, | 917 (25.3%) | 1078 (25.1%) | 31 (7.6%) | 19 (9.0%) |
| Steady average pace, | 1843 (50.8%) | 2142 (49.9%) | 206 (50.2%) | 110 (52.4%) |
| Brisk pace) | 863 (23.8%) | 1067 (24.8%) | 173 (42.2%) | 81 (38.6%) |
| Smoking Status, % | | | | |
| (Never, | 2315 (63.9%) | 1674 (39.0%) | 243 (59.3%) | 121 (57.6%) |
| Former, | 1012 (27.9%) | 2490 (58.0%) | 161 (39.3%) | 82 (39.0%) |
| Current) | 285 (7.9%) | 130 (3.0%) | 6 (1.5%) | 7 (3.3%) |
| Prior Fracture: upper arm/shoulder) | 219 (6.0%) | 50 (1.2%) | 50 (12.2%) | 22 (10.5%) |
| Prior Fracture: wrist | 393 (10.8%) | 69 (1.6%) | 107 (26.1%) | 24 (11.4%) |

**Table 2.** Performance (mean ± SD) on UKB 410 Females with 5 fracture cases (target cohort). SOF was used as the source cohort.

| MMD | CORAL | GRL / DANN | MLP (Baseline) | AUC (mean ± SD) | Performance | ΔAUC | p-value |
|---|---|---|---|---|---|---|---|
| | | | ✓ (Base line) | 0.81 ± 0.02 | Accuracy: 0.80 ± 0.019<br>Sensitivity: 1.00 ± 0.00<br>Specificity: 0.79 ± 0.02<br>Precision: 0.05 ± 0.00<br>F1-score: 0.90 ± 0.01 | – | – |
| ✓ | | | ✓ | 0.92 ± 0.03 | Accuracy: 0.90± 0.02<br>Sensitivity: 0.50 ± 0.00<br>Specificity: 0.99 ± 0.02<br>Precision: 0.58 ± 0.11<br>F1-score: 0.49 ± 0.10 | 0.11 | **0.0001*** |
| | ✓ | | ✓ | 0.81 ± 0.02 | Accuracy: 0.79 ± 0.02<br>Sensitivity: 1.00 ± 0.00<br>Specificity: 0.79 ± 0.02<br>Precision: 0.05 ± 0.00<br>F1--score : 0.09 ± 0.01 | 0.00 | 0.8466 (ns) |
| | | ✓ | ✓ | 0.95 ± 0.02 | Accuracy: 0.96 ± 0.02<br>Sensitivity: 0.08 ± 0.00<br>Specificity: 0.96 ± 0.02<br>Precision: 0.23 ± 0.03<br>F1-score: 0.33 ± 0.04 | 0.14 | **0.0012** |
| ✓ | ✓ | | ✓ | 0.93 ± 0.03 | Accuracy: 0.98 ± 0.02<br>Sensitivity: 0.50 ± 0.00<br>Specificity: 0.98 ± 0.02<br>Precision: 0.43 ± 0.05<br>F1-score: 0.40 ± 0.05 | 0.12 | **0.0003*** |
| ✓ | | ✓ | ✓ | 0.94 ± 0.03 | Accuracy: 0.94 ± 0.029<br>Sensitivity: 0.90 ± 0.00<br>Specificity: 0.94 ± 0.03<br>Precision: 0.14 ± 0.02<br>F1-score: 0.24 ± 0.04 | 0.13 | **0.0016** |
| | ✓ | ✓ | ✓ | 0.95 ± 0.01 | Accuracy: 0.96 ± 0.01<br>Sensitivity: 0.80 ± 0.00<br>Specificity: 0.96 ± 0.01<br>Precision: 0.16 ± 0.02<br>F1-score: 0.27 ± 0.03 | 0.14 | **0.0002*** |
| ✓ | ✓ | ✓ | ✓ | **0.95 ± 0.02** | Accuracy: **0.96 ± 0.02**<br>Sensitivity: **0.90 ± 0.00**<br>Specificity: **0.96 ± 0.02**<br>Precision: **0.19 ± 0.02**<br>F1-score: **0.30 ± 0.03** | **0.14** | **0.0001*** |

**Table 3.** Performance (mean ± SD) on UKB 210 Males with 3 fracture cases (target cohort). MrOS was used as the source cohort.

| **MMD** | **CORAL** | **GRL/DANN** | **MLP (Baseline)** | **AUC (mean ± SD)** | **Performance (3 decimals)** | **ΔAUC** | **p-value** |
|---|---|---|---|---|---|---|---|
| | | | ✓ (Baseline) | 0.71 ± 0.05 | Accuracy: 0.71 ± 0.04<br>Sensitivity: 1.00 ± 0.00<br>Specificity: 0.71 ± 0.05<br>Precision: 0.03 ± 0.06<br>F1-score: 0.06 ± 0.01 | – | – |
| ✓ | | | ✓ | 0.87 ± 0.05 | Accuracy: 0.87 ± 0.048<br>Sensitivity: 1.00 ± 0.00<br>Specificity: 0.868 ± 0.048<br>Precision: 0.074 ± 0.019<br>F1-score: 0.138 ± 0.034 | 0.16 | 0.0001* |
| | ✓ | | ✓ | 0.71 ± 0.05 | Accuracy: 0.72 ± 0.05<br>Sensitivity: 1.00 ± 0.00<br>Specificity: 0.71 ± 0.05<br>Precision: 0.03 ± 0.01<br>F1-score: 0.07 ± 0.01 | 0.00 | 0.4512 (ns) |
| | | ✓ | ✓ | 0.80 ± 0.08 | Accuracy: 0.80 ± 0.08<br>Sensitivity: 1.00 ± 0.00<br>Specificity: 0.80 ± 0.08<br>Precision: 0.06 ± 0.03<br>F1-score: 0.10 ± 0.05 | 0.09 | 0.0027 |
| ✓ | ✓ | | ✓ | 0.86 ± 0.02 | Accuracy: 0.86 ± 0.02<br>Sensitivity: 1.00 ± 0.00<br>Specificity: 0.86 ± 0.02<br>Precision: 0.07 ± 0.01<br>F1-score: 0.13 ± 0.02 | 0.15 | <0.0001* |
| ✓ | | ✓ | ✓ | 0.88 ± 0.02 | Accuracy: 0.88 ± 0.020<br>Sensitivity: 1.00 ± 0.00<br>Specificity: 0.88 ± 0.02<br>Precision: 0.06 ± 0.03<br>F1-score: 0.14 ± 0.02 | 0.17 | <0.0001* |
| | ✓ | ✓ | ✓ | 0.79 ± 0.08 | Accuracy: 0.79 ± 0.08<br>Sensitivity: 1.00 ± 0.00<br>Specificity: 0.79 ± 0.08<br>PrecisionF1: 0.09 ± 0.03 | 0.08 | 0.0088 |
| ✓ | ✓ | ✓ | ✓ (Full) | **0.88 ± 0.04** | Accuracy: 0.88 ± 0.04<br>Sensitivity: 1.00 ± 0.00<br>Specificity: 0.88 ± 0.04<br>Precision: 0.08 ± 0.03<br>F1-score: 0.15 ± 0.05 | 0.17 | <0.0001* |

**Figure**

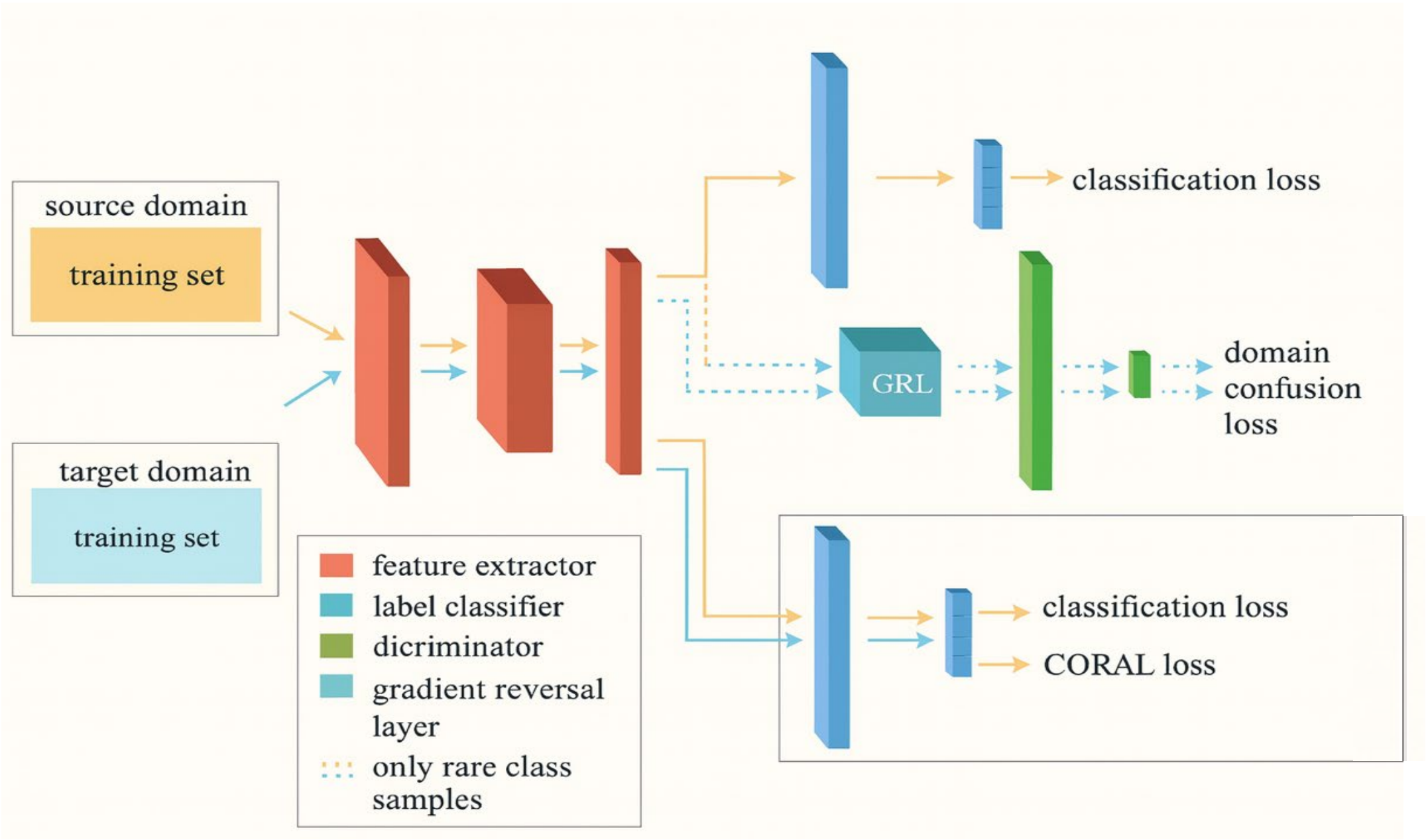

**Figure** 1 Flowchart of GRL + CORAL Model.